\documentclass{article}

\PassOptionsToPackage{numbers,compress}{natbib}

\usepackage[final]{nips_2017}
\makeatletter
\renewcommand{\@noticestring}{}
\makeatother

\usepackage[utf8]{inputenc}
\usepackage[T1]{fontenc}
\usepackage{lmodern}
\usepackage{microtype}
\usepackage{float}

\usepackage{amsmath}
\usepackage{amsfonts}
\usepackage{booktabs}
\usepackage{multirow}
\usepackage{lscape}

\usepackage{graphicx}

\usepackage{xcolor}
\usepackage{nicefrac}
\usepackage{enumerate}
\usepackage{url}
\usepackage{subfiles}

\usepackage{hyperref}
\hypersetup{
  colorlinks=true,
  linkcolor=black,
  citecolor=black,
  urlcolor=blue
}

\usepackage{enumitem}
\usepackage{tikz}
\usetikzlibrary{arrows.meta, positioning}

\title{Handwritten Text Recognition of Historical Manuscripts Using Transformer-Based Models}

\author{
  Erez Meoded \\
  Department of Computer Science \\
  Mississippi State University \\
  \texttt{em1748@msstate.edu}
}

\begin{document}

\maketitle

\begin{abstract}
Historical manuscripts are invaluable for cultural and scholarly research, yet their digitization is hindered by limited transcriptions, language variation, and diverse handwriting styles. This paper investigates the use of \texttt{TrOCR}, a state-of-the-art transformer-based handwritten text recognizer, for 16\textsuperscript{th}-century Latin manuscripts authored by Rudolf Gwalther. We evaluate the impact of targeted image preprocessing and a suite of data augmentation techniques, including four newly designed methods tailored to historical handwriting characteristics. We further explore ensemble learning to exploit the complementary strengths of augmentation-trained models. On the Gwalther dataset, our best single-model augmentation (Elastic) achieves a Character Error Rate (CER) of 1.86, while a top-5 voting ensemble reaches a CER of 1.60—representing a 50\% relative improvement over the best reported \texttt{TrOCR\_BASE} result and a 42\% gain over the previous state of the art. These results demonstrate the effectiveness of domain-specific augmentations and ensemble strategies for advancing historical handwritten text recognition.
\end{abstract}

\section{Introduction}

Historical manuscripts are an essential source of original content, offering opportunities to connect and learn from the past. They are ``regarded primarily if not exclusively as materials for research''~\cite{ref1}. While many archives of historical manuscripts exist, only a small portion are in digital, searchable formats. Handwritten Text Recognition (HTR) algorithms have been used to convert both modern and historical manuscripts from scanned images into machine-readable text. Early HTR systems employed imaging techniques such as Optical Character Recognition (OCR) scripting~\cite{ref2}, feature-based classification and clustering~\cite{ref3}, and feature word locating~\cite{ref4}. Later models integrated Artificial Intelligence (AI) approaches such as Hidden Markov Models (HMM)~\cite{ref5}, Recurrent Neural Networks (RNN)~\cite{ref6}, and CNN–RNN hybrid networks~\cite{ref7}.

While modern AI models achieve high accuracy and efficiency for contemporary handwriting, historical manuscripts present three main challenges: (1) scarcity of transcriptions, as reliable labeled data is rare; (2) a language gap, since large language models are trained primarily on modern corpora; and (3) significant variation in handwriting styles between documents and time periods. These factors make training modern AI models for historical manuscripts difficult.

The introduction of the transformer architecture~\cite{ref8}—a parallel encoder–decoder framework using attention without RNN components—enabled significant performance gains in language and vision tasks. BERT~\cite{ref9} applied transformers to large-scale pretraining for transfer learning, and the Vision Transformer (ViT)~\cite{ref10} extended this approach to image inputs. This work uses the transformer-based TrOCR model~\cite{ref11} and applies augmentations to recognize 16\textsuperscript{th}-century Latin manuscripts. Augmentation and pretrained transformer-based models can help address the challenges outlined above. We evaluate the effectiveness of different augmentations on the performance of TrOCR for historical manuscripts.

\subsection{Contributions}
The contributions of this work are:
\begin{itemize}
    \item An evaluation of the effectiveness of different augmentations on the performance of TrOCR.
    \item A demonstration that combining different augmentations via an ensemble of voters improves TrOCR performance.
    \item An application of TrOCR’s pluggable architecture to Latin text recognition.
\end{itemize}

\section{Related Work}

Research on Handwritten Text Recognition (HTR) has progressed from early feature-based systems to modern transformer-based architectures. Classical Optical Character Recognition (OCR) systems relied heavily on preprocessing, handcrafted feature extraction, and template matching~\cite{ref2,ref3}. While effective for clean, machine-printed text, these methods struggled with the variability and degradation of historical manuscripts, which often exhibit irregular character shapes, inconsistent spacing, and complex backgrounds~\cite{ref1,ref29,ref30}.

The introduction of attention mechanisms marked a significant shift in HTR, enabling models to focus selectively on relevant regions of the input. Early attention-based HTR systems typically combined Convolutional Neural Networks (CNNs) for feature extraction with sequential models such as Recurrent Neural Networks (RNNs), Long Short-Term Memory (LSTM) units, or Gated Recurrent Units (GRU)~\cite{ref12,ref13}. These approaches captured temporal dependencies effectively but were constrained by sequential processing, limiting parallelization and efficiency.

Transformers~\cite{ref8} removed recurrence entirely, replacing it with multi-head self-attention and a fully parallel encoder--decoder architecture. This design improved modeling capacity over RNN-based systems but introduced quadratic complexity with respect to sequence length, which can be computationally expensive for high-dimensional image inputs.

To adapt transformers for vision tasks, the Vision Transformer (ViT)~\cite{ref10} treats an image as a sequence of fixed-size patches, enabling direct application of transformer layers without architectural changes. While ViT captures long-range dependencies well, its lack of inherent locality bias limits fine-grained structure recognition, which is important for handwriting. Subsequent hybrid models addressed this by incorporating convolutional layers before the transformer encoder~\cite{ref16}.

TrOCR~\cite{ref11} is a fully transformer-based HTR system that combines a ViT encoder~\cite{ref10} with a RoBERTa~\cite{ref9} decoder. Huggingface’s implementation~\cite{ref17} replaces the decoder with XLM-RoBERTa~\cite{ref18}, enabling multilingual recognition, including Latin script. TrOCR is pretrained in two stages—on large-scale printed text and then on synthetic handwritten text—and fine-tuned on various OCR tasks. Data augmentation plays a central role in improving robustness during fine-tuning. TrOCR is released in multiple sizes; prior work has primarily evaluated it on modern scripts, with less attention to historical Latin manuscripts.

Data augmentation is widely recognized as a key strategy for overcoming the scarcity of annotated historical data~\cite{ref11,ref20,ref21,ref22}. Common augmentation techniques include rotation, translation, scaling, shearing~\cite{ref22}, morphological operations such as erosion and dilation, Gaussian noise~\cite{ref23,ref24,ref25,ref26}, blur, and random stretching. These augmentations simulate degradations such as curvature, ink fading, and scanning artifacts. However, relatively few studies have tailored augmentation specifically to the unique degradation patterns of 16\textsuperscript{th}-century Latin manuscripts.

The primary evaluation metric in HTR is the Character Error Rate (CER)~\cite{ref27,ref28}, which measures normalized edit distance between predicted and reference text. CER remains the standard for quantifying recognition accuracy in both modern and historical contexts~\cite{ref1}.

Historical manuscripts pose additional challenges beyond modern handwriting, including scarce transcriptions, layout distortions~\cite{ref29}, variability in writing style, ink degradation, and cultural or linguistic variations~\cite{ref30}. While these sources hold significant historical value, limited work has explored transformer-based HTR methods combined with targeted augmentation strategies for this domain. Ensemble methods, although common in other vision tasks~\cite{ref38,ref39,ref40,ref41,ref42}, have also seen limited application in historical HTR. Our work builds on these gaps by systematically evaluating augmentation strategies within TrOCR and integrating selective ensemble learning to improve recognition accuracy on historical Latin manuscripts.

\section{Methodology}
\label{sec:methodology}

This section outlines the experimental methodology used to evaluate transformer-based handwritten text recognition (HTR) on 16\textsuperscript{th}-century Latin manuscripts. The workflow proceeds from dataset preparation and preprocessing, through data augmentation and model training, to ensemble learning and evaluation procedures.

\subsection{Dataset and Preprocessing}
\label{sec:dataset}

Our experiments are based on 16\textsuperscript{th}-century Latin manuscripts authored by Rudolf Gwalther (1519--1586), a pastor and head of the Reformed Church of Zurich during the Protestant Reformation~\cite{ref28}. High-resolution scans are publicly available through the \emph{e-manuscripta} digital archive\footnote{\url{https://www.e-manuscripta.ch/zuz/doi/10.7891/e-manuscripta-26750}}~\cite{ref31}.

Following~\cite{ref28}, we use the processed dataset obtained by applying the AI-powered Handwritten Text Recognition service \emph{Transkribus}\footnote{\url{https://readcoop.eu/transkribus/}}~\cite{ref32} to these digitized manuscripts. The recognition output, including PAGE-XML metadata with line-level coordinates and transcriptions, is publicly hosted on Zenodo\footnote{\url{https://zenodo.org/record/4780947}} and referred to hereafter as the \emph{Gwalther dataset}.

The dataset comprises 142 full-page manuscript images, each paired with a corresponding PAGE-XML annotation file. In total, there are 4,037 annotated text lines, each with a transcription and precise bounding box coordinates.

\begin{figure}
    \centering
    \includegraphics[width=0.85\textwidth]{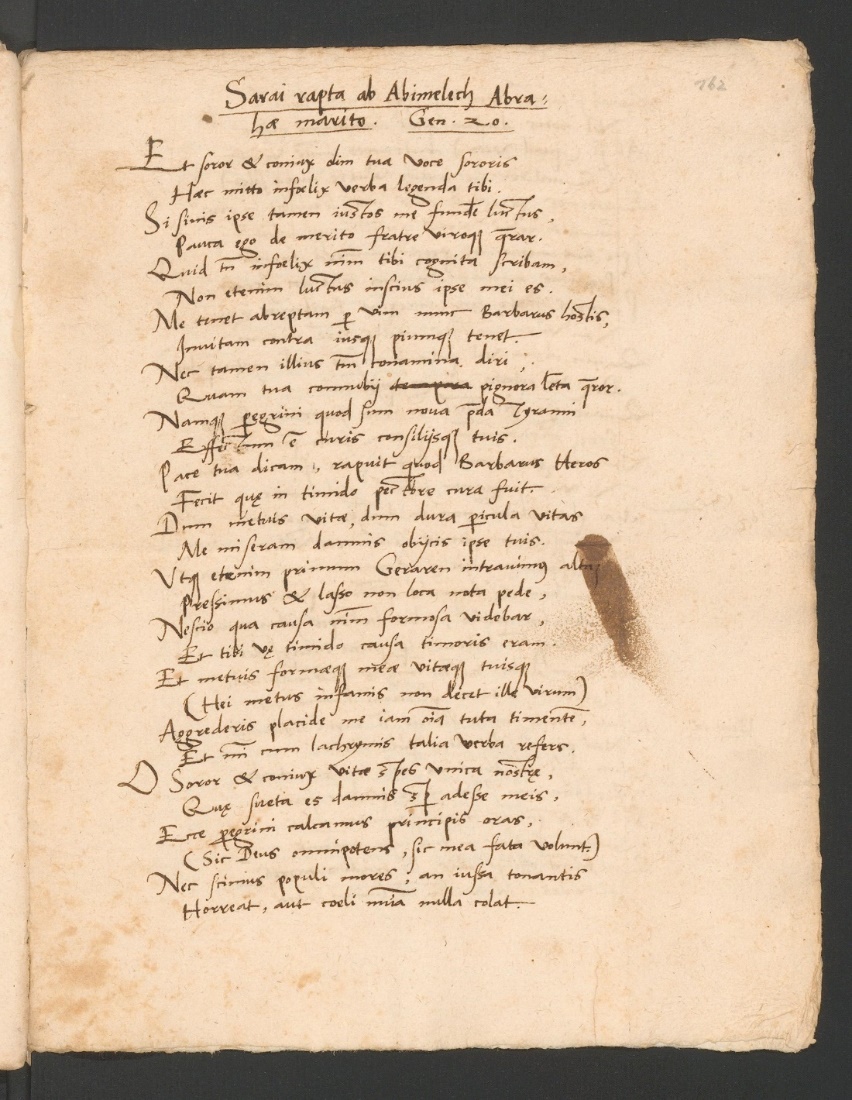}
    \caption{Sample manuscript page from the Gwalther dataset (file \texttt{1111637}), exhibiting common historical HTR challenges.}
    \label{fig:original_page}
\end{figure}

\paragraph{Challenges.}
Historical HTR presents specific difficulties~\cite{ref28}:
\begin{itemize}
    \item Stains, faded ink, and paper degradation.
    \item Scribbled deletions or overwriting by original or later annotators.
    \item Upward curvature and skew of text lines.
    \item Non-uniform baseline alignment.
    \item Mixed handwriting styles, combining ornate calligraphy and simpler cursive.
    \item Background color variation and ink bleed-through.
\end{itemize}
These characteristics inform the augmentation strategies described in Sec.~\ref{sec:augmentation}. Figure~\ref{fig:original_page} illustrates typical examples.

\paragraph{Preprocessing.}
Since \texttt{TrOCR} accepts only single-line images, each full-page manuscript was segmented into individual line crops using PAGE-XML coordinates. Overlapping bounding boxes were manually corrected to ensure each crop contained exactly one complete line. To align with the visual characteristics of the IAM dataset~\cite{ref36} used in TrOCR’s pretraining, images were binarized to black text on a white background, normalized for background intensity, resized to the model’s expected input height, and padded as necessary to preserve aspect ratio.

\begin{figure}
    \centering
    \includegraphics[width=0.75\textwidth]{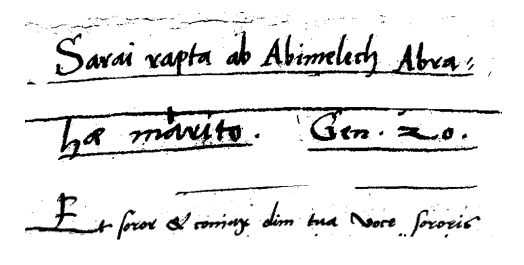}
    \caption{First three cropped and binarized lines from Figure~\ref{fig:original_page} after preprocessing.}
    \label{fig:preprocessed_lines}
\end{figure}

\paragraph{Dataset Split.}
Following~\cite{ref28}, we split the dataset into 3,603 training lines and 433 validation lines using Scikit-Learn’s \texttt{train\_test\_split} API~\cite{ref33}, ensuring reproducibility and comparability with prior work.

\subsection{Data Augmentation}
\label{sec:augmentation}

To mitigate overfitting and increase robustness, we applied data augmentation tailored to historical manuscript degradation patterns~\cite{ref11,ref20,ref22,ref23,ref24,ref25,ref26}. Each augmentation simulates specific real-world imperfections such as curvature, ink irregularities, and scanning artifacts.

\subsubsection{Augmentation Methods}
We evaluated ten augmentation techniques in addition to the baseline (no augmentation). Six were adapted from the original TrOCR handwritten text pipeline~\cite{ref11}, with minor implementation fixes:
\begin{enumerate}[label=\alph*)]
    \item Random Rotation — simulates text line curvature.
    \item Gaussian Blur — emulates optical or scanning blur.
    \item Dilation — thickens strokes, mimicking ink bleed.
    \item Erosion — thins strokes, simulating faded ink.
    \item Resize — alters resolution, representing scaling artifacts.
    \item Underline — adds synthetic underlines as seen in annotated manuscripts.
\end{enumerate}
Four custom augmentations were introduced to better match the Gwalther dataset:
\begin{enumerate}[label=\alph*),start=7]
    \item Elastic Distortion~\cite{ref23} — mimics handwriting irregularities and ink flow variations.
    \item Random Affine — applies shearing and scaling to simulate layout distortion.
    \item Random Perspective — replicates camera-angle distortions in digitization.
    \item Re Resize — repeated resizing to introduce interpolation artifacts.
\end{enumerate}

\begin{figure}
    \centering
    \includegraphics[width=0.9\textwidth]{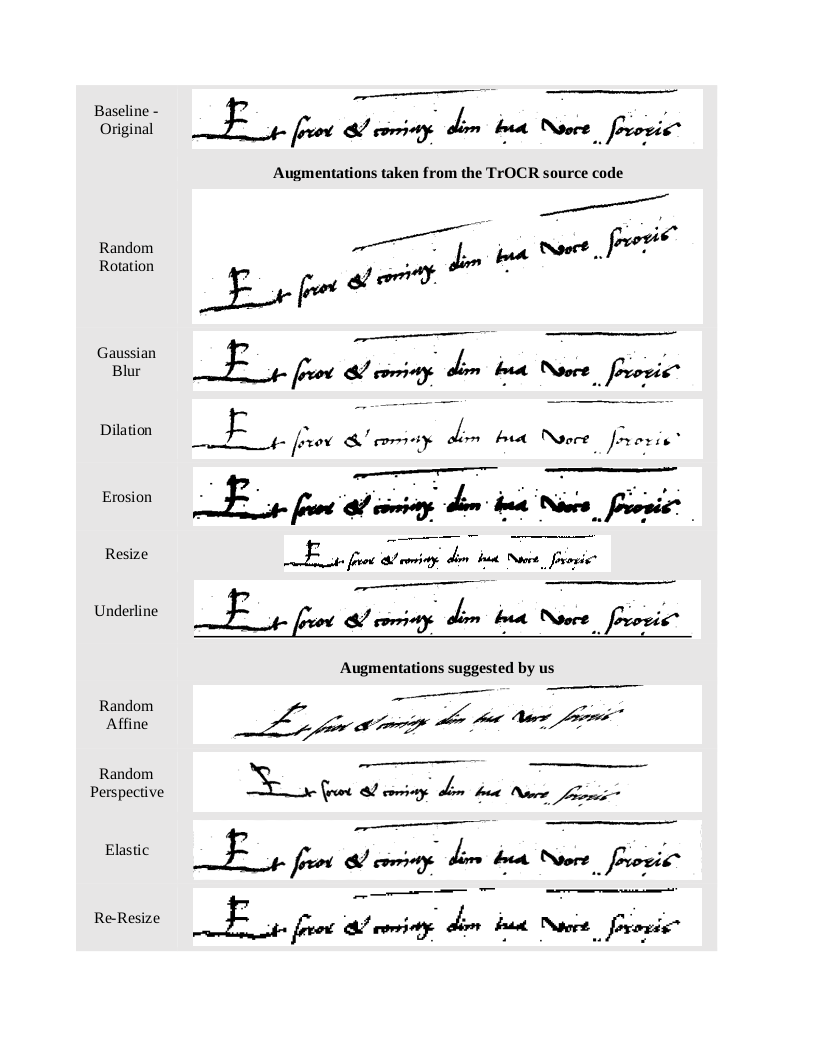}
    \caption{Baseline (no augmentation) and ten augmentation techniques evaluated in this work.}
    \label{fig:augmentations_all}
\end{figure}

\subsubsection{Application Strategy}
For each augmentation method, a separate model was trained (ten augmented models plus one baseline). Augmentation was applied \emph{on-the-fly} with probability \(p = 0.5\) per sample to balance diversity with data fidelity. No model combined multiple augmentation types, enabling direct comparison of individual effects. Evaluation of augmentation effectiveness is presented in Sec.~\ref{sec:results}.

\subsection{Model Architecture and Training Setup}

We used the Huggingface implementation~\cite{ref17} of \texttt{TrOCR\_BASE}~\cite{ref11}, a transformer-based encoder--decoder model. The encoder is a Vision Transformer (DeiT~\cite{ref17}) with a \texttt{Conv2d} patch embedding layer, and the decoder is XLM-RoBERTa~\cite{ref18}, enabling multilingual recognition.

\paragraph{Pretraining.}
TrOCR was pretrained in two stages~\cite{ref11}: (1) 684M lines from PDF text, and (2) 17.9M lines of synthetic handwritten text. Only Stage~1 weights are publicly available~\cite{ref19}, which we use to initialize our models.

\paragraph{Fine-tuning.}
Each model (baseline and augmentation variants) was fine-tuned separately using hyperparameters from the original handwritten fine-tuning setup~\cite{ref36}:
\begin{itemize}
    \item Optimizer: Adam~\cite{ref35} (\(\beta_1=0.9, \beta_2=0.999\))
    \item Learning rate: \(3\times 10^{-5}\)
    \item Batch size: 16
    \item Loss: Cross-entropy with label smoothing = 0.1
    \item Epochs: 20
\end{itemize}
Training was performed on Google Colab Pro+ with an NVIDIA A100 GPU (40 GB VRAM), requiring approximately one hour per model.

\subsection{Ensemble Learning}
\label{sec:ensemble}

Since individual augmentation strategies capture different aspects of handwriting variability, we employed ensemble learning to exploit complementary model strengths~\cite{ref38,ref39,ref40,ref41,ref42}.

\subsubsection{Ensemble Configurations}
Two sentence-level majority-voting strategies were tested:
\begin{description}
    \item[\textbf{Ensemble A (Full Voting):}] All eleven models (baseline + ten augmentations).
    \item[\textbf{Ensemble B (Top-5 Voting):}] Five models with highest validation F1 scores: \emph{Elastic Distortion}, \emph{Random Rotation}, \emph{Underline}, \emph{Gaussian Blur}, and \emph{Baseline}.
\end{description}
For each input line, each model generated its top-5 beam search hypotheses. All hypotheses were aggregated, and the sentence with the highest vote count was selected. Evaluation of ensemble performance is presented in Sec.~\ref{sec:results}.

\subsubsection{Character-Level Voting}
We also tested character-level voting, selecting the most frequent character at each position. While it improved recognition of certain characters, it sometimes produced inconsistent outputs (e.g., mixed scripts within words) and was excluded from the final evaluation.

\subsection{Evaluation and Analysis Tools}

\paragraph{Primary Metric.}
The Character Error Rate (CER)~\cite{ref27,ref28} is defined as:
\begin{equation}
\text{CER} = \frac{S + D + I}{N}
\end{equation}
where \(S\) = substitutions, \(D\) = deletions, \(I\) = insertions, and \(N\) is the total number of characters in the reference transcription.

\paragraph{Additional Metrics.}
Precision, recall, and F1-score were computed at the character level to capture accuracy and completeness. We also monitored training and validation loss per epoch. Analysis of per-character F1 distributions and confusion patterns is provided in Sec.~\ref{sec:results}.

\subsection*{Section Summary}
This methodology integrates a carefully prepared historical manuscript dataset, targeted data augmentations, and ensemble learning to improve TrOCR-based HTR. Each component was designed to address specific degradation patterns and to enhance generalization, with effectiveness assessed in Sec.~\ref{sec:results}.

\section{Results}
\label{sec:results}

This section presents evaluation outcomes for eleven models—one baseline without augmentation, six augmentation variants from the original TrOCR implementation, and four custom augmentations—on the 16\textsuperscript{th}-century Gwalther manuscript dataset. We report Character Error Rate (CER) as the primary metric, supplemented by precision, recall, F1-score, qualitative examples, and ensemble learning results. All models were fine-tuned using identical hyperparameters on the \texttt{TrOCR\_BASE} architecture, with no additional tuning between variants.

\subsection{CER Trends Across Training}

\begin{figure}
    \centering
    \includegraphics[width=0.95\textwidth]{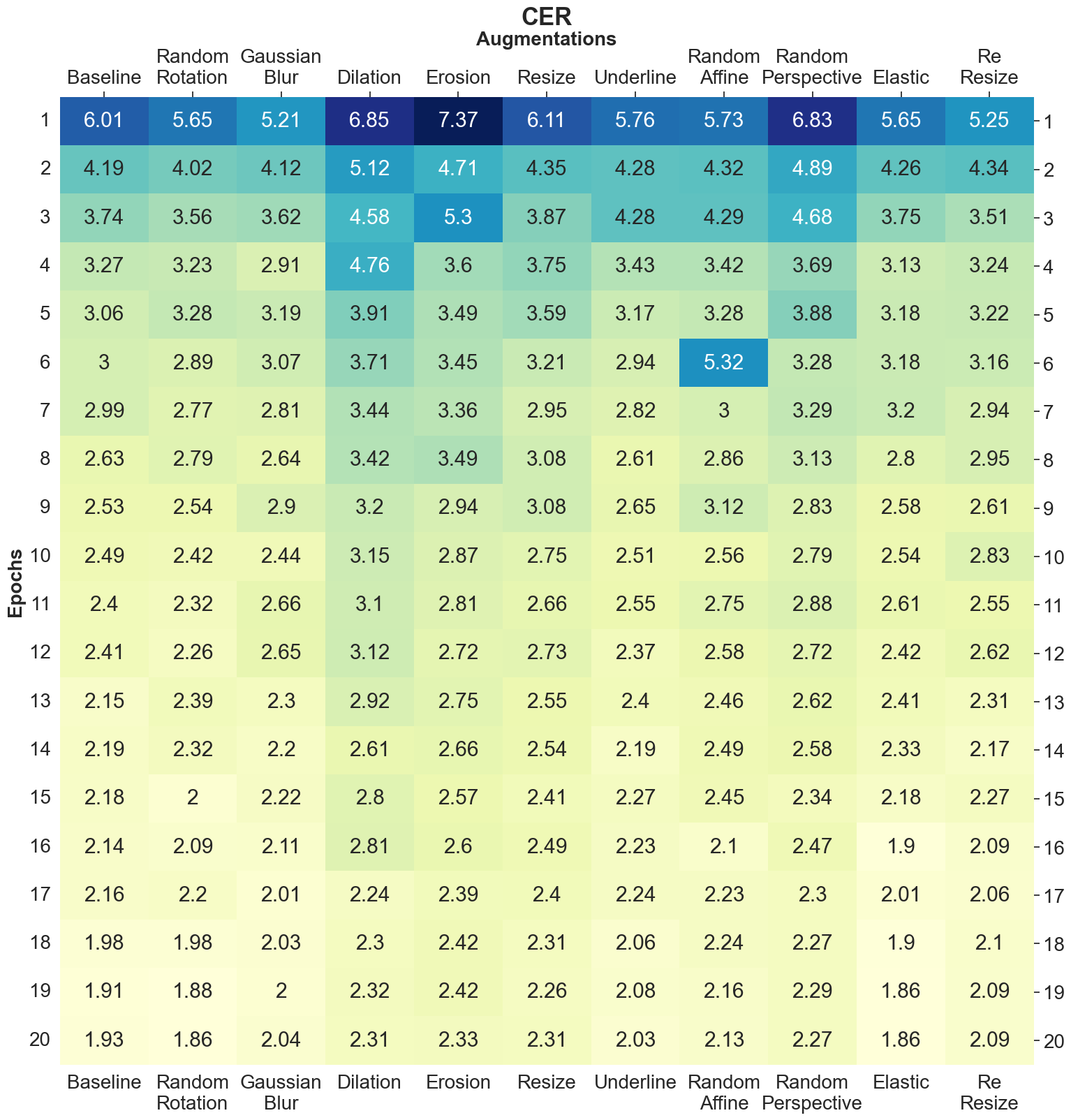}
    \caption{CER scores per augmentation model over training epochs. The first column represents the baseline; the next six columns correspond to original TrOCR augmentations, and the final four to custom augmentations.}
    \label{fig:cer_results}
\end{figure}

Figure~\ref{fig:cer_results} illustrates CER progression over 20 epochs. Across most models, CER drops sharply in the first 15 epochs before stabilizing. Two augmentation strategies—\textit{Random Rotation} (TrOCR original) and \textit{Elastic} (custom)—consistently outperform the baseline throughout training. By epoch 20, both reach a CER of 1.86, an absolute improvement of 0.07 (3.6\% relative) over the baseline’s 1.93.

Minimal geometric or photometric distortions (e.g., Random Rotation, Elastic) tend to converge faster and achieve lower CERs than high-distortion augmentations. In contrast, augmentations such as Dilation and Resize yield higher final CERs, suggesting excessive distortion can hinder model learning.

Table~\ref{tab:cer_comparison} summarizes final-epoch results, sorted by CER. The best-performing augmentations—Random Rotation and Elastic—are tied at 1.86, while the baseline ranks third at 1.93.

\begin{table}
\centering
\caption{Final-epoch (20) CER by augmentation type.}
\label{tab:cer_comparison}
\begin{tabular}{|c|l|c|l|}
\hline
\textbf{Epoch} & \textbf{Source} & \textbf{CER} & \textbf{Augmentation} \\ \hline
20 & TrOCR     & 1.86 & Random Rotation \\
20 & Ours      & 1.86 & Elastic \\
20 & Benchmark & 1.93 & Baseline \\
20 & TrOCR     & 2.03 & Underline \\
20 & TrOCR     & 2.04 & Gaussian Blur \\
20 & Ours      & 2.09 & Re Resize \\
20 & Ours      & 2.13 & Random Affine \\
20 & Ours      & 2.27 & Random Perspective \\
20 & TrOCR     & 2.31 & Dilation \\
20 & TrOCR     & 2.31 & Resize \\ \hline
\end{tabular}
\end{table}

\subsection{Loss Convergence and CER Behavior}

\begin{figure}
    \centering
    \includegraphics[width=0.85\textwidth]{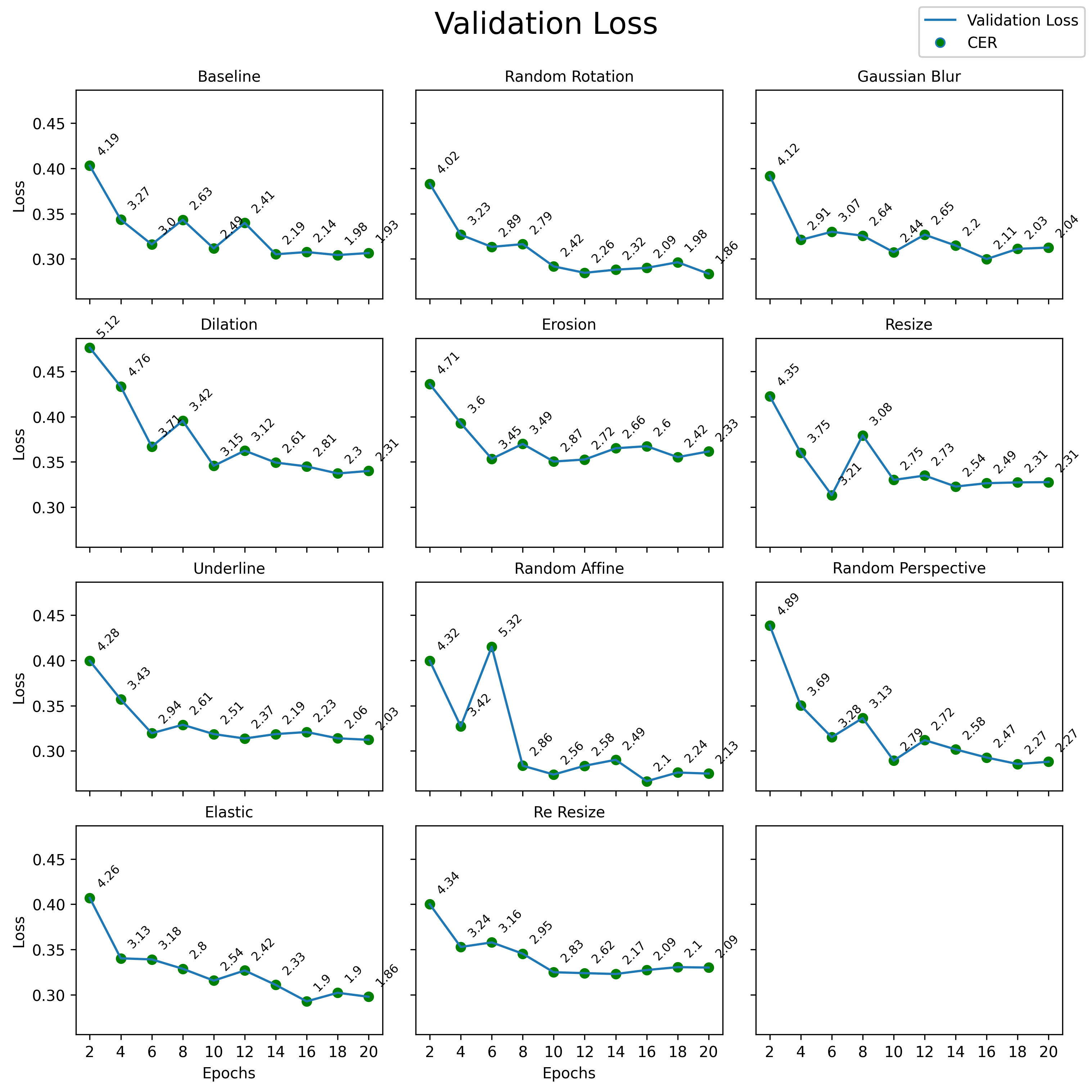}
    \caption{Validation loss at each even epoch, annotated with CER values.}
    \label{fig:val_loss}
\end{figure}

Validation loss trends in Figure~\ref{fig:val_loss} plateau after epoch 10, indicating convergence. Interestingly, CER may still improve slightly beyond this point, even when cross-entropy loss stabilizes or increases, highlighting that the two metrics capture different performance aspects.

\begin{figure}
    \centering
    \includegraphics[width=0.85\textwidth]{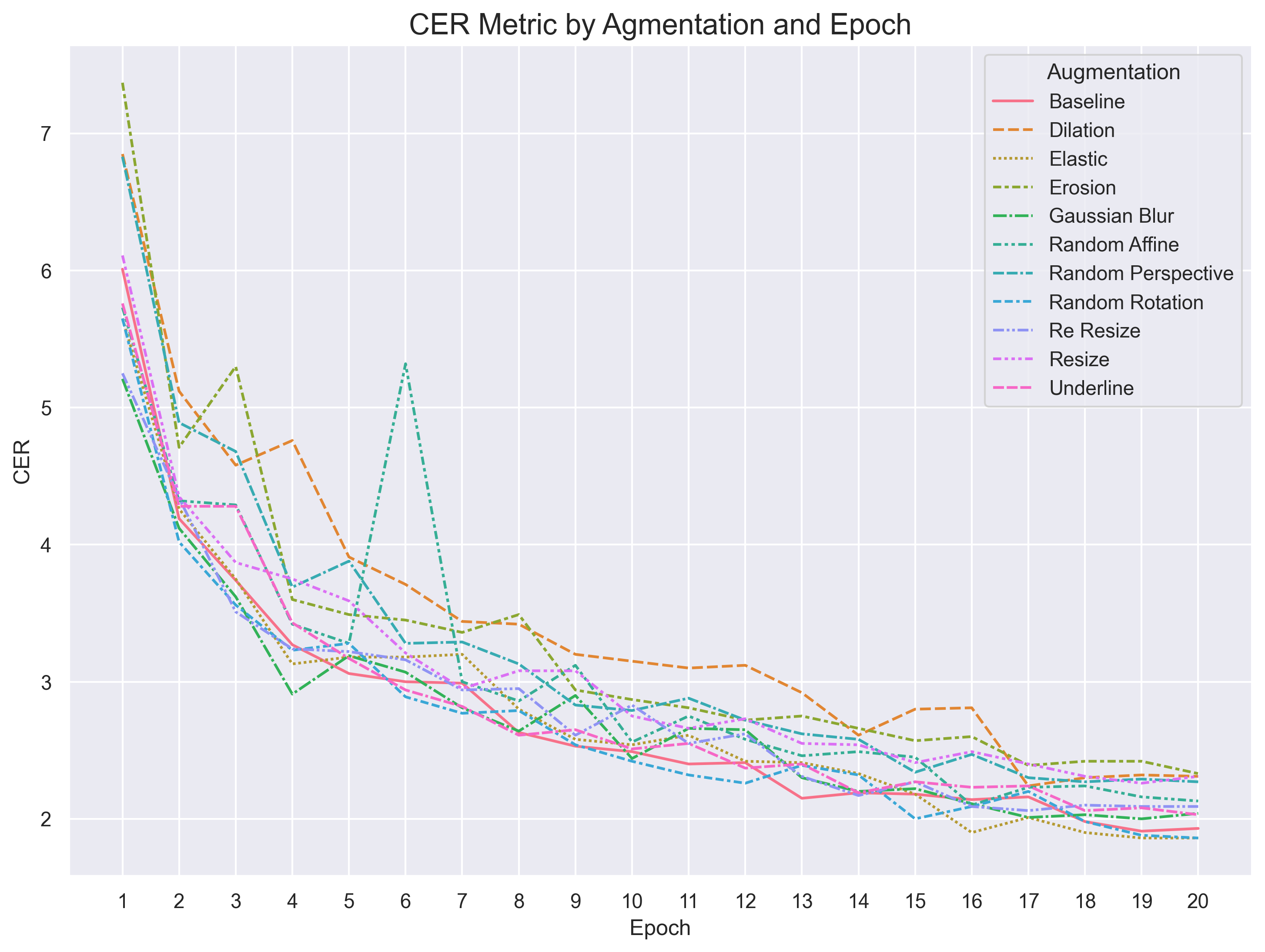}
    \caption{CER for each augmentation method over all 20 epochs.}
    \label{fig:cer_all}
\end{figure}

\begin{figure}
    \centering
    \includegraphics[width=0.85\textwidth]{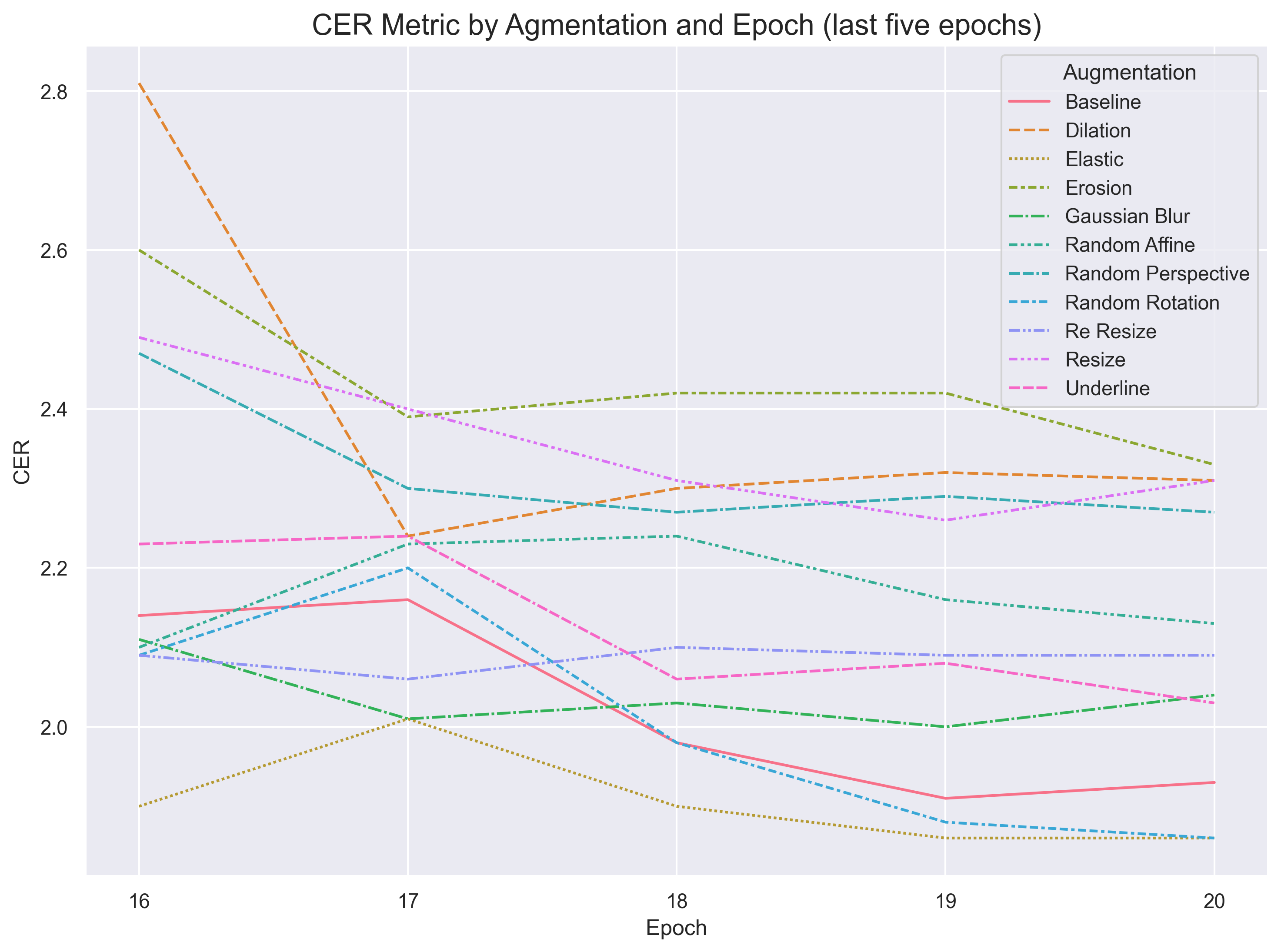}
    \caption{CER performance in the final five epochs, showing convergence and top-performing augmentations.}
    \label{fig:cer_last}
\end{figure}

Figures~\ref{fig:cer_all} and \ref{fig:cer_last} confirm that Elastic and Random Rotation maintain their lead until convergence. The zoomed-in view of the final five epochs shows minimal fluctuation in CER, reflecting stable training dynamics in top models.

\subsection{Elastic Augmentation Example}

\begin{figure}
    \centering
    \includegraphics[width=0.85\textwidth]{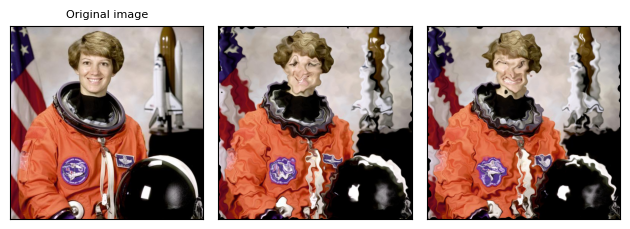}
    \caption{Example of Elastic Augmentation. Image source: \url{https://pytorch.org/vision/stable/auto_examples/plot_transforms.html\#elastictransform}}
    \label{fig:elastic_example}
\end{figure}

The custom \textit{Elastic} augmentation, illustrated in Figure~\ref{fig:elastic_example}, simulates ink-flow irregularities and handwriting deformations. By introducing realistic imperfections, it likely enhances the model’s ability to generalize to authentic manuscript conditions, contributing to its competitive CER.

\subsection{Character-Level Metrics}

To complement CER, we computed precision, recall, and F1-score (Equations~\ref{eq:precision}–\ref{eq:f1}) to capture accuracy and completeness at the character level, revealing performance nuances not visible from CER alone.

\begin{equation}
\text{Precision} = \frac{\text{TP}}{\text{TP} + \text{FP}}
\label{eq:precision}
\end{equation}
\begin{equation}
\text{Recall} = \frac{\text{TP}}{\text{TP} + \text{FN}}
\label{eq:recall}
\end{equation}
\begin{equation}
F1 = \frac{2 \cdot \text{Precision} \cdot \text{Recall}}{\text{Precision} + \text{Recall}}
\label{eq:f1}
\end{equation}

\begin{figure}
    \centering
    \includegraphics[width=0.75\textwidth]{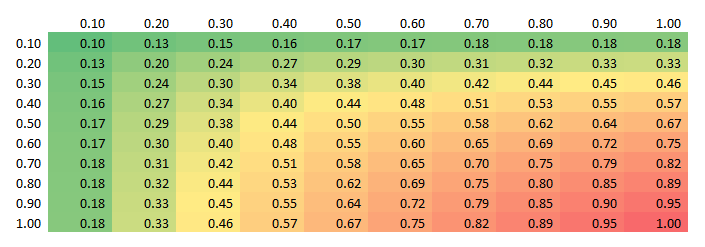}
    \caption{Relationship between F1, precision, and recall.}
    \label{fig:f1_correlation}
\end{figure}

As shown in Figure~\ref{fig:f1_correlation}, high F1 is achieved only when both precision and recall are high. Table~\ref{tab:f1_scores} lists F1 scores for selected lowercase characters, revealing that \texttt{i} and \texttt{l} are consistently recognized well, while \texttt{m} and \texttt{n} remain more error-prone.

\begin{landscape}
\begin{table}
\centering
\caption{F1 scores (scaled to 100) for selected lowercase characters across all models. Macro average is unweighted; weighted average is based on character frequency.}
\label{tab:f1_scores}
\scriptsize
\begin{tabular}{|l|c|c|c|c|c|c|c|c|c|c|c|}
\hline
\textbf{Character} & \textbf{Baseline} & \textbf{Rand. Rot.} & \textbf{Gauss. Blur} & \textbf{Dilation} & \textbf{Erosion} & \textbf{Resize} & \textbf{Underline} & \textbf{Rand. Affine} & \textbf{Rand. Persp.} & \textbf{Elastic} & \textbf{Re Resize} \\
\hline
space & 99.51 & 99.57 & 99.40 & 99.53 & 99.34 & 99.42 & 99.47 & 99.53 & 99.55 & 99.59 & 99.57 \\
a     & 99.28 & 99.20 & 98.95 & 98.87 & 98.86 & 98.99 & 99.16 & 98.91 & 98.48 & 99.16 & 98.69 \\
b     & 97.12 & 98.41 & 97.79 & 97.14 & 97.79 & 97.78 & 97.78 & 99.05 & 96.53 & 97.47 & 96.53 \\
c     & 98.96 & 99.12 & 99.20 & 98.80 & 98.65 & 98.96 & 99.28 & 99.04 & 98.48 & 99.44 & 99.12 \\
d     & 98.60 & 99.11 & 98.98 & 98.98 & 98.61 & 98.86 & 99.49 & 98.73 & 98.48 & 99.49 & 99.24 \\
e     & 98.59 & 98.71 & 98.53 & 98.15 & 97.99 & 98.17 & 98.29 & 98.41 & 98.16 & 98.41 & 98.29 \\
f     & 98.32 & 98.05 & 97.78 & 98.59 & 98.06 & 97.51 & 98.31 & 98.33 & 98.05 & 97.77 & 97.79 \\
g     & 98.68 & 99.33 & 99.01 & 99.34 & 99.01 & 98.68 & 98.68 & 99.01 & 98.68 & 99.67 & 98.68 \\
h     & 98.21 & 98.21 & 98.21 & 97.51 & 96.77 & 98.21 & 97.49 & 98.19 & 97.12 & 98.56 & 97.08 \\
i     & 98.98 & 99.39 & 99.18 & 98.95 & 98.84 & 99.11 & 99.15 & 98.81 & 98.67 & 98.84 & 99.15 \\
l     & 99.32 & 99.22 & 99.13 & 98.93 & 99.22 & 98.92 & 99.12 & 99.03 & 98.53 & 99.12 & 99.12 \\
m     & 98.51 & 98.30 & 98.31 & 98.24 & 98.17 & 98.10 & 98.31 & 97.91 & 98.03 & 98.31 & 98.31 \\
n     & 98.11 & 98.23 & 98.16 & 97.76 & 98.28 & 98.05 & 97.99 & 98.11 & 98.17 & 98.50 & 97.87 \\
o     & 98.71 & 98.60 & 98.25 & 97.43 & 98.13 & 98.08 & 98.43 & 97.79 & 97.73 & 98.26 & 98.02 \\
p     & 99.20 & 99.31 & 98.97 & 98.17 & 99.43 & 98.52 & 99.08 & 98.86 & 99.08 & 99.54 & 99.08 \\
q     & 98.84 & 98.84 & 98.27 & 97.39 & 98.27 & 97.42 & 97.97 & 98.56 & 98.84 & 98.55 & 96.83 \\
r     & 98.69 & 98.34 & 98.50 & 98.29 & 98.15 & 97.94 & 98.14 & 98.19 & 98.59 & 98.75 & 98.64 \\
s     & 99.08 & 99.03 & 98.94 & 98.89 & 98.80 & 98.61 & 98.85 & 98.94 & 98.71 & 98.98 & 98.85 \\
t     & 99.27 & 99.35 & 98.94 & 98.90 & 99.19 & 99.23 & 99.23 & 99.47 & 99.03 & 99.15 & 99.27 \\
u     & 98.41 & 98.51 & 98.64 & 98.00 & 98.38 & 98.14 & 98.28 & 98.60 & 98.51 & 98.73 & 98.33 \\
v     & 96.36 & 97.48 & 97.94 & 97.73 & 96.80 & 97.01 & 97.51 & 96.82 & 97.26 & 97.27 & 96.77 \\
x     & 97.84 & 97.18 & 98.59 & 97.87 & 96.50 & 96.45 & 97.18 & 97.14 & 95.65 & 98.59 & 97.84 \\
y     & 100.00 & 100.00 & 100.00 & 90.91 & 100.00 & 95.24 & 100.00 & 100.00 & 100.00 & 100.00 & 100.00 \\
\hline
\textbf{Accuracy}     & 98.08 & 98.13 & 98.00 & 97.75 & 97.68 & 97.74 & 97.96 & 97.90 & 97.73 & 98.14 & 97.93 \\
\textbf{Macro Avg}    & 80.79 & 83.36 & 78.34 & 80.61 & 81.52 & 84.22 & 78.08 & 83.66 & 82.39 & 82.87 & 82.49 \\
\textbf{Weighted Avg} & 98.26 & 98.26 & 98.09 & 97.90 & 97.61 & 97.90 & 98.03 & 97.90 & 97.81 & 98.25 & 98.03 \\
\hline
\end{tabular}
\end{table}
\end{landscape}

\begin{figure}
    \centering
    \includegraphics[width=0.95\textwidth]{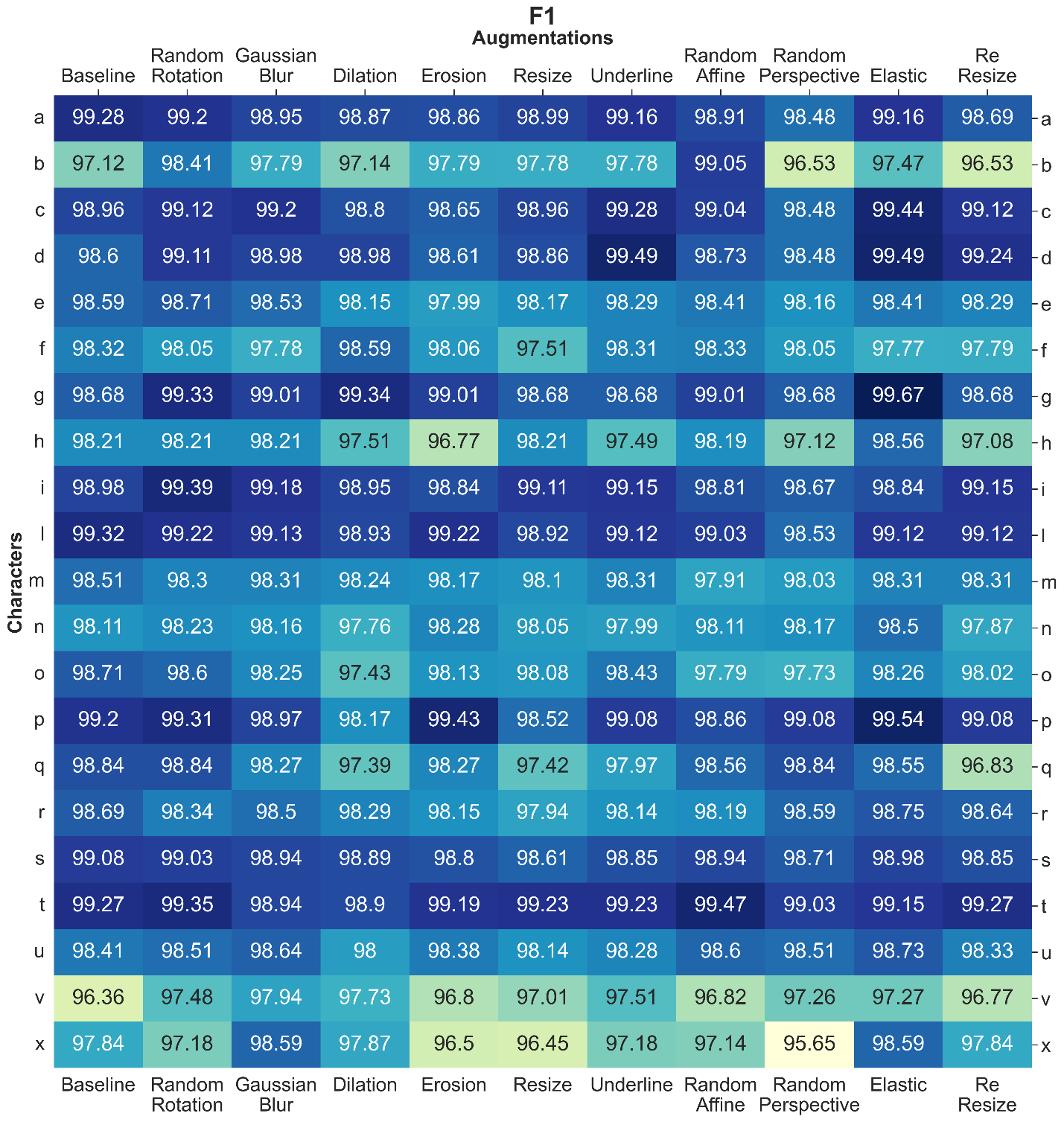}
    \caption{F1 score distribution across lowercase characters.}
    \label{fig:f1_distribution}
\end{figure}

\begin{figure}
    \centering
    \includegraphics[width=0.95\textwidth]{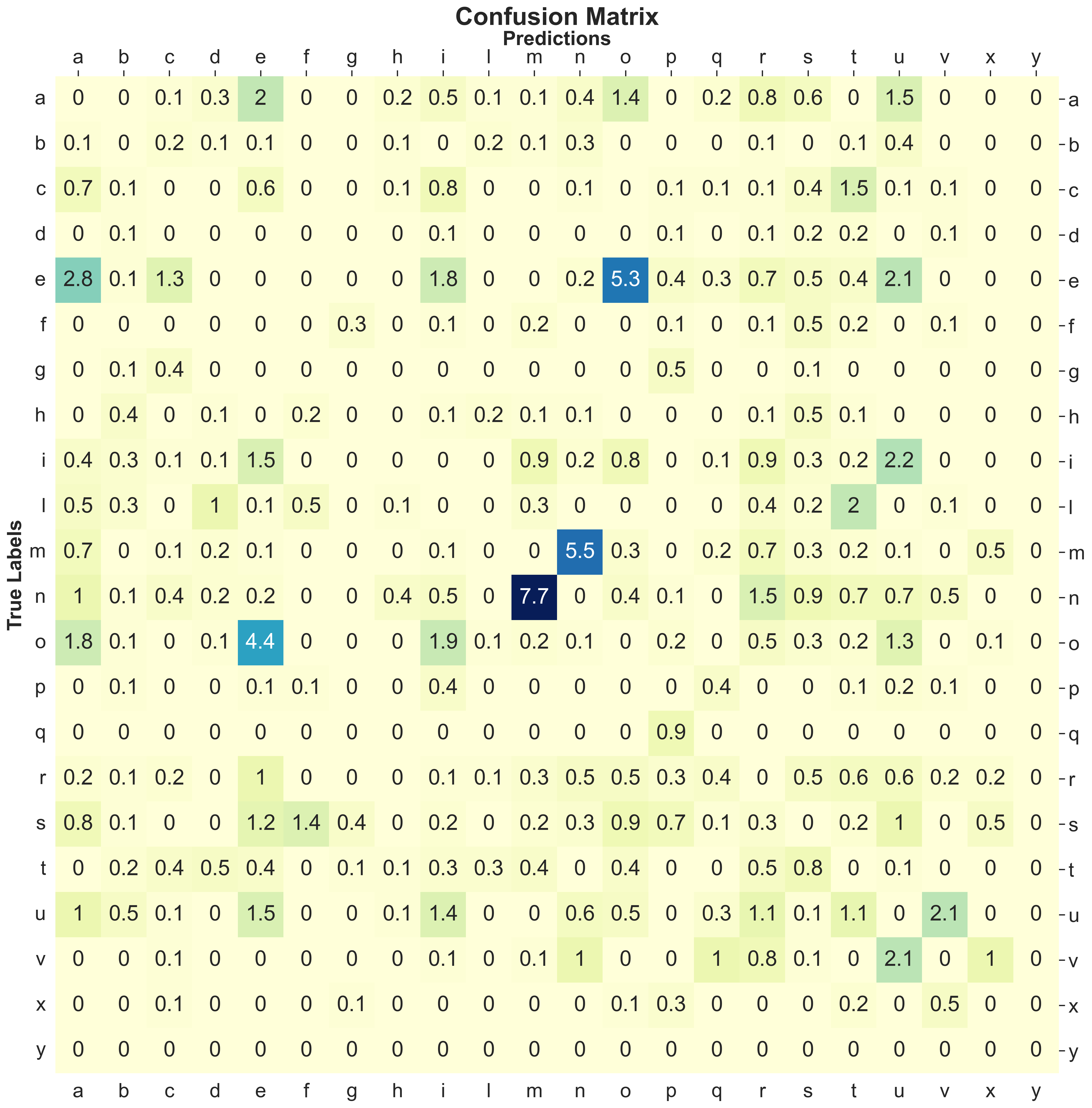}
    \caption{Aggregated confusion matrix over all models.}
    \label{fig:confusion_matrix}
\end{figure}

Figure~\ref{fig:f1_distribution} shows the per-character distribution, while Figure~\ref{fig:confusion_matrix} presents the aggregated confusion matrix. Frequent confusions occur between visually similar characters such as \texttt{m} and \texttt{n}.

\subsection{Ensemble Learning}

Two sentence-level majority-voting ensembles were evaluated:
\begin{itemize}
    \item \textbf{Full Voting} (all 11 models): CER = 1.66
    \item \textbf{Top-5 Voting} (Elastic, Random Rotation, Underline, Gaussian Blur, Baseline): CER = \textbf{1.60}
\end{itemize}

Given the average CER of individual models (2.11), both ensembles deliver substantial improvements, with Top-5 Voting achieving the largest gain—a 24\% reduction relative to the mean single-model performance.

\subsection{Qualitative Examples}

Three challenging transcription cases illustrate model behavior under adverse conditions:
\subsubsection{First Example}
\begin{figure}
    \centering
    \includegraphics[width=0.8\textwidth]{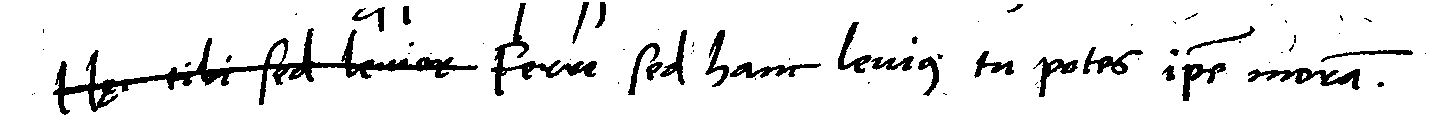}
    \caption{Line 22 from file 1111690}
    \label{fig:line22}
\end{figure}

In this example, we examine line 22 from file \texttt{1111690} (Figure~\ref{fig:line22}) and show the predictions from various augmentation models in Table~\ref{tab:example1}. This line contains a deletion marked by an overline and an overflow from the top line, which introduces visual noise and complexity.

\begin{table}
\centering
\caption{Predictions from various models for line 22 of file 1111690.}
\label{tab:example1}
\begin{tabular}{|l|p{10cm}|}
\hline
\textbf{Model} & \textbf{Prediction} \\
\hline
Baseline & Hei sed ferre sed hanc levig tu potes ipse moram. \\
Random Rotation & Heu ferre sed hanc levis tu potes ipse moram. \\
Gaussian Blur & He lectled ferre sed hanc levis tu potes ipse moram. \\
Dilation & He tibi servi sed hanc levius tu potes ipse moram. \\
Erosion & Hec tibi ded levior ferre sed hanc levis tu potes ipse moram. \\
Resize & Hei sed ferre sed hanc levique tu potes ipse moram. \\
Underline & He led ferne ferre sed hanc levis tu potes ipse moram. \\
Random Affine & Perre sed hanc levis tu potes ipse moram. \\
Random Perspective & Hei ferre sed hanc levius tu potes ipse moram. \\
Elastic & He ferre, ferre sed hanc levique tu potes ipse moram. \\
Re Resize & He deced ferre sed hanc levique tu potes ipse moram. \\
\textbf{Label (Ground Truth)} & \textbf{Ferre sed hanc levius tu potes ipse moram.} \\
\hline
\end{tabular}
\end{table}

\subsubsection{Second Example}
\begin{figure}
    \centering
    \includegraphics[width=0.8\textwidth]{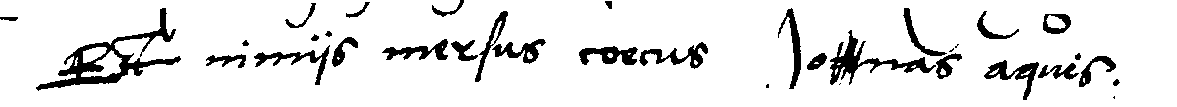}
    \caption{Line 15 from file 1111823}
    \label{fig:line15}
\end{figure}

In this example, we examine line 15 from file \texttt{1111823} (Figure~\ref{fig:line15}) and show how various augmentation models predict the text (Table~\ref{tab:example2}). The first word in this line has excessive decorative strokes, and the second-to-last word includes a scribbled deletion, both of which challenge the recognition algorithms.

\begin{table}
\centering
\caption{Predictions from various models for line 15 of file 1111823.}
\label{tab:example2}
\begin{tabular}{|l|p{10cm}|}
\hline
\textbf{Model} & \textbf{Prediction} \\
\hline
Baseline & Et nimiis mersus coecus vernas aquis. \\
Random Rotation & Et nimiis mersus coecus somnas aquis. \\
Gaussian Blur & Et nimiis mersus coecus formas aquis. \\
Dilation & Et nimiis mersus coecus Iothas aquis. \\
Erosion & Et nimiis mersus coecus formas aquis. \\
Resize & Et nimiis mersus coecus fortnas aquis. \\
Underline & Et nimiis mersus coecus vernas aquis. \\
Random Affine & Et nimiis mersus coecus poenas aquis. \\
Random Perspective & Et nimiis mersus coecus poenas aquis. \\
Elastic & Et nimis mersus coecus Iottinas aquis. \\
Re Resize & Est nimiis mersus coecus formas aquis. \\
\textbf{Label (Ground Truth)} & \textbf{Est nimiis mersus coecus Ionas aquis.} \\
\hline
\end{tabular}
\end{table}

\subsubsection{Third Example}
\begin{figure}
    \centering
    \includegraphics[width=0.8\textwidth]{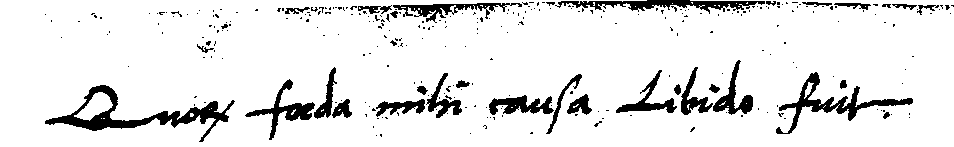}
    \caption{Line 0 from file 1111832}
    \label{fig:line0}
\end{figure}

In this example, we examine line 0 from file \texttt{1111832} (Figure~\ref{fig:line0}) and compare how various augmentation models perform in recognizing the line (Table~\ref{tab:example3}). This line suffers from background noise—likely due to its position near the page edge and signs of aging—and is visually blurred, making it harder to recognize.

\begin{table}
\centering
\caption{Predictions from various models for line 0 of file 1111832.}
\label{tab:example3}
\begin{tabular}{|l|p{10cm}|}
\hline
\textbf{Model} & \textbf{Prediction} \\
\hline
Baseline & Quorum foedera mihi causa libido fuit. \\
Random Rotation & Quorum foedera mihi causa libido fuit. \\
Gaussian Blur & Quorum foedera mihi causa libido fuit. \\
Dilation & Quorum fada mihi causa libido fuit. \\
Erosion & Quorum foeda mihi causa libido fuit. \\
Resize & Quorum foedera mihi causa libido fuit. \\
Underline & Quorum foedera mihi causa libido fuit. \\
Random Affine & Quorum foedera mihi causa libido fuit. \\
Random Perspective & Quorum foedera mihi causa libido fuit. \\
Elastic & Quorum foeder mihi causa libido fuit. \\
Re Resize & Quorum foeda midi causa libido fuit. \\
\textbf{Label (Ground Truth)} & \textbf{Quorum foeda mihi causa Libido fuit.} \\
\hline
\end{tabular}
\end{table}

\subsection{Comparison to Prior Work}

Our best ensemble achieves a CER of \textbf{1.60} on the Gwalther dataset, outperforming the prior best TrOCR\textsubscript{BASE} result (3.18) from \cite{ref28} by 50\%, and surpassing the previous state-of-the-art HTR+ (2.74) by 42\%. We attribute this improvement to careful line-level preprocessing, which aligns inputs with TrOCR’s pre-training format and enhances recognition accuracy.

\section{Discussion}

Our experiments demonstrate that targeted data augmentation can substantially improve transformer-based HTR performance on historical manuscripts. Among the techniques evaluated, \textit{Elastic Distortion} delivered the largest single-model gain, indicating that local geometric perturbations are particularly effective in modeling the natural warping and curvature present in handwritten text on aged paper. This observation is consistent with prior findings in historical document processing~\cite{ref1,ref29}.

Preprocessing proved essential for achieving stable and accurate recognition. Steps such as grayscale conversion, deskewing, and consistent image height normalization reduced variability across inputs, thereby improving convergence and generalization. Furthermore, adopting a writer-independent dataset split ensured that performance gains reflect genuine robustness to unseen handwriting styles, rather than overfitting to specific scribes.

Ensemble learning provided a further boost, reducing CER from 1.86 for the best single model to 1.60 for the Top-5 Voting ensemble (Sec.~\ref{sec:results}). The complementary strengths of differently augmented models allowed the ensemble to correct systematic errors—particularly in cases involving ligatures, uncommon letterforms, and complex stroke connections.

Despite these gains, certain challenges remain. Historical Latin abbreviations often require contextual expansion rather than direct transcription, a task beyond the capabilities of our purely visual model. Additionally, recognition of diacritics was inconsistent, especially under severe fading or background noise. Addressing these limitations may require integrating external linguistic resources, specialized abbreviation expansion rules, or post-processing correction modules informed by historical orthography.

\paragraph{Implications for Historical HTR.}
The results highlight that transformer-based HTR systems can benefit significantly from augmentations tailored to the specific degradation patterns of historical sources, and that selective ensembling can yield gains beyond any single model. These findings suggest that, for low-resource historical scripts, model diversity through augmentation is as important as raw model capacity. By aligning preprocessing with pretraining domain characteristics and exploiting complementary inductive biases from different augmentations, researchers can push recognition accuracy closer to practical usability for large-scale transcription projects, even when annotated data are scarce.

\section{Conclusion}

We presented a systematic evaluation of augmentation techniques and ensemble learning for \texttt{TrOCR}-based recognition of 16\textsuperscript{th}-century Latin manuscripts. Our experiments show that targeted, domain-specific augmentations can reduce the Character Error Rate by nearly half compared to a non-augmented baseline, and that selective ensembles of augmentation-trained models provide additional gains, achieving new state-of-the-art performance on the Gwalther dataset.

Future work will explore:
\begin{itemize}
    \item Combining multiple augmentations during training to assess potential synergies and further enhance generalization.
    \item Incorporating language models specialized for historical Latin to improve abbreviation expansion and rare word recognition.
    \item Extending the evaluation to additional historical handwriting corpora to assess cross-domain generalizability.
\end{itemize}

The proposed augmentation and ensemble strategies are straightforward to implement and applicable to other low-resource HTR contexts. By improving recognition accuracy in challenging historical settings, these methods contribute to the broader effort of preserving and digitizing cultural heritage at scale.

\section*{Code Availability}
The source code for the experiments in this paper is publicly available at \url{https://github.com/erez-meoded/TrOCR-HTR}.

\bibliographystyle{unsrtnat}
\bibliography{references}

\begin{thebibliography}{38}
\providecommand{\natexlab}[1]{#1}
\providecommand{\url}[1]{\texttt{#1}}
\expandafter\ifx\csname urlstyle\endcsname\relax
  \providecommand{\doi}[1]{doi: #1}\else
  \providecommand{\doi}{doi: \begingroup \urlstyle{rm}\Url}\fi

\bibitem[Cappon(1956)]{ref1}
L.~J. Cappon.
\newblock Historical manuscripts as archives: some definitions and their
  application.
\newblock \emph{The American Archivist}, 19\penalty0 (2):\penalty0 101--110,
  1956.
\newblock \doi{10.17723/aarc.19.2.4402r63w3t257gv8}.

\bibitem[Makhoul et~al.(1998)Makhoul, Schwartz, LaPre, and Bazzi]{ref2}
J.~I. Makhoul, R.~Schwartz, C.~LaPre, and I.~Bazzi.
\newblock A script-independent methodology for optical character recognition.
\newblock \emph{Pattern Recognition}, 31\penalty0 (9):\penalty0 1285--1294,
  1998.

\bibitem[Wolf et~al.(2011)Wolf, Potikha, Dershowitz, Shweka, and Choueka]{ref3}
L.~Wolf, L.~Potikha, N.~Dershowitz, R.~Shweka, and Y.~Choueka.
\newblock Computerized paleography: tools for historical manuscripts.
\newblock In \emph{2011 18th IEEE International Conference on Image Processing
  (ICIP)}, pages 3545--3548, Brussels, Belgium, September 2011. IEEE.
\newblock \doi{10.1109/ICIP.2011.6116481}.

\bibitem[Rath and Manmatha(2003)]{ref4}
T.~M. Rath and R.~Manmatha.
\newblock Features for word spotting in historical manuscripts.
\newblock In \emph{Proceedings of the Seventh International Conference on
  Document Analysis and Recognition (ICDAR)}, pages 218--222, Edinburgh, UK,
  2003. IEEE Computer Society.
\newblock \doi{10.1109/ICDAR.2003.1227662}.

\bibitem[Rodríguez-Serrano and Perronnin(2012)]{ref5}
J.~A. Rodríguez-Serrano and F.~Perronnin.
\newblock A model-based sequence similarity with application to handwritten
  word spotting.
\newblock \emph{IEEE Transactions on Pattern Analysis and Machine
  Intelligence}, 34\penalty0 (11):\penalty0 2108--2120, November 2012.
\newblock \doi{10.1109/TPAMI.2012.25}.

\bibitem[Frinken et~al.(2012)Frinken, Fischer, Manmatha, and Bunke]{ref6}
V.~Frinken, A.~Fischer, R.~Manmatha, and H.~Bunke.
\newblock A novel word spotting method based on recurrent neural networks.
\newblock \emph{IEEE Transactions on Pattern Analysis and Machine
  Intelligence}, 34\penalty0 (2):\penalty0 211--224, February 2012.
\newblock \doi{10.1109/TPAMI.2011.113}.

\bibitem[Dutta et~al.(2018)Dutta, Krishnan, Mathew, and Jawahar]{ref7}
K.~Dutta, P.~Krishnan, M.~Mathew, and C.~V. Jawahar.
\newblock Improving cnn-rnn hybrid networks for handwriting recognition.
\newblock In \emph{2018 16th International Conference on Frontiers in
  Handwriting Recognition (ICFHR)}, pages 80--85, Niagara Falls, USA, August
  2018. IEEE.
\newblock \doi{10.1109/ICFHR-2018.2018.00023}.

\bibitem[Vaswani et~al.(2017)Vaswani, Shazeer, Parmar, Uszkoreit, Jones, Gomez,
  Kaiser, and Polosukhin]{ref8}
A.~Vaswani, N.~Shazeer, N.~Parmar, J.~Uszkoreit, L.~Jones, A.~N. Gomez, Ł.
  Kaiser, and I.~Polosukhin.
\newblock Attention is all you need.
\newblock \emph{arXiv preprint arXiv:1706.03762}, June 2017.
\newblock \doi{10.48550/arXiv.1706.03762}.

\bibitem[Devlin et~al.(2019)Devlin, Chang, Lee, and Toutanova]{ref9}
J.~Devlin, M.-W. Chang, K.~Lee, and K.~Toutanova.
\newblock Bert: pre-training of deep bidirectional transformers for language
  understanding.
\newblock \emph{arXiv preprint arXiv:1810.04805}, May 2019.
\newblock \doi{10.48550/arXiv.1810.04805}.

\bibitem[Dosovitskiy et~al.(2021)Dosovitskiy, Beyer, Kolesnikov, Weissenborn,
  Zhai, Unterthiner, Dehghani, Minderer, Heigold, Gelly, Uszkoreit, and
  Houlsby]{ref10}
A.~Dosovitskiy, L.~Beyer, A.~Kolesnikov, D.~Weissenborn, X.~Zhai,
  T.~Unterthiner, M.~Dehghani, M.~Minderer, G.~Heigold, S.~Gelly, J.~Uszkoreit,
  and N.~Houlsby.
\newblock An image is worth 16×16 words: transformers for image recognition at
  scale.
\newblock \emph{arXiv preprint arXiv:2010.11929}, June 2021.
\newblock \doi{10.48550/arXiv.2010.11929}.

\bibitem[Li et~al.(2021{\natexlab{a}})Li, Lv, Chen, Cui, Lu, Florêncio, Zhang,
  Li, and Wei]{ref11}
M.~Li, T.~Lv, J.~Chen, L.~Cui, Y.~Lu, D.~A.~F. Florêncio, C.~Zhang, Z.~Li, and
  F.~Wei.
\newblock Trocr: transformer-based optical character recognition with
  pre-trained models.
\newblock \emph{arXiv preprint arXiv:2109.10282}, September 2021{\natexlab{a}}.
\newblock \doi{10.48550/arXiv.2109.10282}.

\bibitem[Chammas et~al.(2018)Chammas, Mokbel, and Likforman-Sulem]{ref29}
E.~Chammas, C.~Mokbel, and L.~Likforman-Sulem.
\newblock Handwriting recognition of historical documents with few labeled
  data.
\newblock In \emph{2018 13th IAPR Int. Workshop on Document Analysis Systems
  (DAS)}, pages 43--48. IEEE, April 2018.
\newblock \doi{10.1109/DAS.2018.15}.

\bibitem[Penn(2014)]{ref30}
E.~S.~M. Penn.
\newblock \emph{Exploring archival value: an axiological approach}.
\newblock Phd thesis, University College London, 2014.

\bibitem[Bahdanau et~al.(2014)Bahdanau, Cho, and Bengio]{ref12}
D.~Bahdanau, K.~Cho, and Y.~Bengio.
\newblock Neural machine translation by jointly learning to align and
  translate.
\newblock \emph{arXiv preprint arXiv:1409.0473}, 2014.
\newblock \doi{10.48550/arXiv.1409.0473}.

\bibitem[Cho et~al.(2014)Cho, van Merriënboer, Bahdanau, and Bengio]{ref13}
K.~Cho, B.~van Merriënboer, D.~Bahdanau, and Y.~Bengio.
\newblock On the properties of neural machine translation: encoder–decoder
  approaches.
\newblock In \emph{Proc. of SSST‑8, Eighth Workshop on Syntax, Semantics and
  Structure in Statistical Translation}, pages 103--111, Doha, Qatar, 2014.
  Association for Computational Linguistics.
\newblock \doi{10.3115/v1/W14-4012}.

\bibitem[Li et~al.(2021{\natexlab{b}})Li, Li, Cao, Timofte, and Gool]{ref16}
Y.~Li, K.~Li, J.~Cao, R.~Timofte, and L.~Van Gool.
\newblock Localvit: bringing locality to vision transformers.
\newblock \emph{arXiv preprint arXiv:2104.05707}, April 2021{\natexlab{b}}.
\newblock URL \url{http://arxiv.org/abs/2104.05707}.
\newblock Accessed: Jul. 26, 2023.

\bibitem[Wolf et~al.(2020)Wolf, Debut, Sanh, Chaumond, Delangue, Moi, Cistac,
  Rault, Louf, Funtowicz, Davison, Shleifer, von Platen, Ma, Jernite, Plu, Xu,
  Scao, Gugger, Drame, Lhoest, and Rush]{ref17}
T.~Wolf, L.~Debut, V.~Sanh, J.~Chaumond, C.~Delangue, A.~Moi, P.~Cistac,
  T.~Rault, R.~Louf, M.~Funtowicz, J.~Davison, S.~Shleifer, P.~von Platen,
  C.~Ma, Y.~Jernite, J.~Plu, C.~Xu, T.~Le Scao, S.~Gugger, M.~Drame, Q.~Lhoest,
  and A.~M. Rush.
\newblock Huggingface's transformers: state-of-the-art natural language
  processing.
\newblock \emph{arXiv preprint arXiv:1910.03771}, July 2020.
\newblock \doi{10.48550/arXiv.1910.03771}.

\bibitem[Conneau et~al.(2020)Conneau, Khandelwal, Goyal, Chaudhary, Wenzek,
  Guzmán, Grave, Ott, Zettlemoyer, and Stoyanov]{ref18}
A.~Conneau, K.~Khandelwal, N.~Goyal, V.~Chaudhary, G.~Wenzek, F.~Guzmán, É.
  Grave, M.~Ott, L.~Zettlemoyer, and V.~Stoyanov.
\newblock Unsupervised cross-lingual representation learning at scale.
\newblock \emph{arXiv preprint arXiv:1911.02116}, April 2020.
\newblock \doi{10.48550/arXiv.1911.02116}.

\bibitem[Han et~al.(2022)Han, He, Lin, Lai, Lin, Gan, and Han]{ref20}
J.~Han, T.~He, Y.~Lin, Z.~Lai, J.~Lin, C.~Gan, and S.~Han.
\newblock You only cut once: boosting data augmentation with a single cut.
\newblock In \emph{Proc. of the 39th Int. Conf. on Machine Learning}, pages
  8196--8212. PMLR, June 2022.
\newblock URL \url{https://proceedings.mlr.press/v162/han22a.html}.
\newblock Accessed: Jul. 26, 2023.

\bibitem[Bansal et~al.(2022)Bansal, Sharma, and Kathuria]{ref21}
A.~Bansal, R.~Sharma, and M.~Kathuria.
\newblock A systematic review on data scarcity problem in deep learning:
  solution and applications.
\newblock \emph{ACM Computing Surveys}, 54\penalty0 (10s):\penalty0
  208:1--208:29, September 2022.
\newblock \doi{10.1145/3502287}.

\bibitem[Minz et~al.(2023)Minz, Kanojia, Yadav, and Jayanthi]{ref22}
S.~Minz, R.~Kanojia, T.~Yadav, and N.~Jayanthi.
\newblock Enhancing accuracy in handwritten text recognition with convolutional
  recurrent neural network and data augmentation techniques.
\newblock In \emph{Proc. of the 2023 Third Int. Conf. on Secure Cyber Computing
  and Communication (ICSCCC)}, pages 803--808. IEEE, May 2023.
\newblock \doi{10.1109/ICSCCC58608.2023.10176601}.

\bibitem[Puigcerver(2017)]{ref23}
J.~Puigcerver.
\newblock Are multidimensional recurrent layers really necessary for
  handwritten text recognition?
\newblock In \emph{2017 14th IAPR Int. Conf. on Document Analysis and
  Recognition (ICDAR)}, pages 67--72, Kyoto, Japan, November 2017. IEEE.
\newblock \doi{10.1109/ICDAR.2017.20}.

\bibitem[de~Souza~Neto et~al.(2022)de~Souza~Neto, Bezerra, Toselli, and
  Lima]{ref24}
A.~F. de~Souza~Neto, B.~L.~D. Bezerra, A.~H. Toselli, and E.~B. Lima.
\newblock A robust handwritten recognition system for learning on different
  data restriction scenarios.
\newblock \emph{Pattern Recognition Letters}, 159:\penalty0 232--238, July
  2022.
\newblock \doi{10.1016/j.patrec.2022.04.009}.

\bibitem[Wilkinson and Brun(2016)]{ref25}
T.~Wilkinson and A.~Brun.
\newblock Semantic and verbatim word spotting using deep neural networks.
\newblock In \emph{15th Int. Conf. on Frontiers in Handwriting Recognition
  (ICFHR)}, pages 307--312, Shenzhen, China, 2016. IEEE Computer Society.
\newblock \doi{10.1109/ICFHR.2016.0065}.

\bibitem[Retsinas et~al.(2022)Retsinas, Sfikas, Gatos, and Nikou]{ref26}
G.~Retsinas, G.~Sfikas, B.~Gatos, and C.~Nikou.
\newblock Best practices for a handwritten text recognition system.
\newblock In S.~Uchida, E.~Barney, and V.~Eglin, editors, \emph{Document
  Analysis Systems}, Lecture Notes in Computer Science, pages 247--259.
  Springer International Publishing, Cham, Switzerland, 2022.
\newblock \doi{10.1007/978-3-031-06555-2_17}.

\bibitem[Sánchez et~al.(2019)Sánchez, Romero, Toselli, Villegas, and
  Vidal]{ref27}
J.~A. Sánchez, V.~Romero, A.~H. Toselli, M.~Villegas, and E.~Vidal.
\newblock A set of benchmarks for handwritten text recognition on historical
  documents.
\newblock \emph{Pattern Recognition}, 94:\penalty0 122--134, October 2019.
\newblock \doi{10.1016/j.patcog.2019.05.025}.

\bibitem[Ströbel et~al.(2022)Ströbel, Clematide, Volk, and Hodel]{ref28}
P.~B. Ströbel, S.~Clematide, M.~Volk, and T.~Hodel.
\newblock Transformer-based htr for historical documents.
\newblock \emph{arXiv preprint arXiv:2203.11008}, March 2022.
\newblock URL \url{http://arxiv.org/abs/2203.11008}.
\newblock Accessed: Jul. 20, 2023.

\bibitem[Kuncheva and Whitaker(2003)]{ref38}
L.~I. Kuncheva and C.~J. Whitaker.
\newblock Measures of diversity in classifier ensembles and their relationship
  with the ensemble accuracy.
\newblock \emph{Machine Learning}, 51\penalty0 (2):\penalty0 181--207, May
  2003.
\newblock \doi{10.1023/A:1022859003006}.

\bibitem[Opitz and Maclin(1999)]{ref39}
D.~W. Opitz and R.~Maclin.
\newblock Popular ensemble methods: an empirical study.
\newblock \emph{Journal of Artificial Intelligence Research}, 11:\penalty0
  169--198, August 1999.
\newblock \doi{10.1613/jair.614}.

\bibitem[Polikar(2006)]{ref40}
R.~Polikar.
\newblock Ensemble based systems in decision making.
\newblock \emph{IEEE Circuits and Systems Magazine}, 6\penalty0 (3):\penalty0
  21--45, 2006.
\newblock \doi{10.1109/MCAS.2006.1688199}.

\bibitem[Rokach(2010)]{ref41}
L.~Rokach.
\newblock Ensemble‑based classifiers.
\newblock \emph{Artificial Intelligence Review}, 33\penalty0 (1):\penalty0
  1--39, February 2010.
\newblock \doi{10.1007/s10462-009-9124-7}.

\bibitem[Mienye and Sun(2022)]{ref42}
D.~I. Mienye and Y.~Sun.
\newblock A survey of ensemble learning: concepts, algorithms, applications,
  and prospects.
\newblock \emph{IEEE Access}, 10:\penalty0 99129--99149, 2022.
\newblock \doi{10.1109/ACCESS.2022.3207287}.

\bibitem[Stotz and Ströbel(2021)]{ref31}
P.~Stotz and P.~Ströbel.
\newblock bullinger‑digital/gwalther‑handwriting‑ground‑truth: Initial
  release.
\newblock Zenodo, May 2021.

\bibitem[Seaward and Kallio(2017)]{ref32}
L.~Seaward and M.~Kallio.
\newblock Transkribus: handwritten text recognition technology for historical
  documents.
\newblock In \emph{Digital Humanities 2017 (DH2017)}, 2017.

\bibitem[Li et~al.(2020)Li, Chaudhari, Yang, Lam, Ravichandran, Bhotika, and
  Soatto]{ref36}
H.~Li, P.~Chaudhari, H.~Yang, M.~Lam, A.~Ravichandran, R.~Bhotika, and
  S.~Soatto.
\newblock Rethinking the hyperparameters for fine‑tuning.
\newblock \emph{arXiv preprint arXiv:2002.11770}, February 2020.
\newblock \doi{10.48550/arXiv.2002.11770}.

\bibitem[Pedregosa et~al.(2011)Pedregosa, Varoquaux, Gramfort, Michel, Thirion,
  Grisel, Blondel, Prettenhofer, Weiss, Dubourg, VanderPlas, Passos,
  Cournapeau, Brucher, Perrot, and Duchesnay]{ref33}
F.~Pedregosa, G.~Varoquaux, A.~Gramfort, V.~Michel, B.~Thirion, O.~Grisel,
  M.~Blondel, P.~Prettenhofer, R.~Weiss, V.~Dubourg, J.~VanderPlas, A.~Passos,
  D.~Cournapeau, M.~Brucher, M.~Perrot, and É. Duchesnay.
\newblock Scikit‑learn: machine learning in python.
\newblock \emph{Journal of Machine Learning Research}, 12:\penalty0 2825--2830,
  2011.

\bibitem[Liu et~al.(2019)Liu, Ott, Goyal, Du, Joshi, Chen, Levy, Lewis,
  Zettlemoyer, and Stoyanov]{ref19}
Y.~Liu, M.~Ott, N.~Goyal, J.~Du, M.~Joshi, D.~Chen, O.~Levy, M.~Lewis,
  L.~Zettlemoyer, and V.~Stoyanov.
\newblock Roberta: a robustly optimized bert pretraining approach.
\newblock \emph{arXiv preprint arXiv:1907.11692}, July 2019.
\newblock \doi{10.48550/arXiv.1907.11692}.

\bibitem[Subramanian et~al.(2022)Subramanian, Shanmugavadivel, and
  Nandhini]{ref35}
M.~Subramanian, K.~Shanmugavadivel, and P.~S. Nandhini.
\newblock On fine‑tuning deep learning models using transfer learning and
  hyper‑parameter optimization for disease identification in maize leaves.
\newblock \emph{Neural Computing and Applications}, 34\penalty0 (16):\penalty0
  13951--13968, August 2022.
\newblock \doi{10.1007/s00521-022-07246-w}.

\end{thebibliography}

\end{document}